\documentclass[10pt,twocolumn,letterpaper]{article}

\usepackage{cvpr}              % To produce the CAMERA-READY version
\usepackage{graphicx}
\usepackage{times}
\usepackage{epsfig}
\usepackage{amsmath}
\usepackage{amssymb}
\usepackage{booktabs}
\usepackage{multirow}
\usepackage{verbatim}
\usepackage{appendix}
\usepackage{subcaption}
\usepackage{adjustbox}
\usepackage{pifont}
\usepackage{enumitem}
\newcommand{\trans}[1]{{#1}^{\ensuremath{\mathsf{T}}}} % transpose

\usepackage[accsupp]{axessibility} % Improves PDF readability for those with disabilities

\usepackage[pagebackref,breaklinks,colorlinks]{hyperref}

\usepackage[capitalize]{cleveref}
\crefname{section}{Sec.}{Secs.}
\Crefname{section}{Section}{Sections}
\Crefname{table}{Table}{Tables}
\crefname{table}{Tab.}{Tabs.}

\def\cvprPaperID{1003} % *** Enter the CVPR Paper ID here
\def\confName{CVPR}
\def\confYear{2022}

\begin{document}

%%%%%%%%% TITLE - PLEASE UPDATE
\title{Point-to-Voxel Knowledge Distillation for LiDAR Semantic Segmentation}

\author{\textbf{Yuenan Hou}$^{1}$, \textbf{Xinge Zhu}$^{2}$, \textbf{Yuexin Ma}$^{3}$, \textbf{Chen Change Loy}$^{4}$, \textbf{and Yikang Li$^{1}$}\\
$^{1}$Shanghai AI Laboratory $^{2}$The Chinese University of Hong Kong\\
$^{3}$ShanghaiTech University $^{4}$S-Lab, Nanyang Technological University\\
$^{1}$\{houyuenan, liyikang\}@pjlab.org.cn, $^{2}$zhuxinge123@gmail.com,\\
$^{3}$mayuexin@shanghaitech.edu.cn, $^{4}$ccloy@ntu.edu.sg
}

\maketitle

\def\algorithmname{PVD}

%%%%%%%%% ABSTRACT
\begin{abstract}
% !TEX root = ./main.tex

This article addresses the problem of distilling knowledge from a large teacher model to a slim student network for LiDAR semantic segmentation. Directly employing previous distillation approaches yields inferior results due to the intrinsic challenges of point cloud, \ie, sparsity, randomness and varying density. To tackle the aforementioned problems, we propose the Point-to-Voxel Knowledge Distillation (\algorithmname), which transfers the hidden knowledge from both point level and voxel level. Specifically, we first leverage both the pointwise and voxelwise output distillation to complement the sparse supervision signals. Then, to better exploit the structural information, we divide the whole point cloud into several supervoxels and design a difficulty-aware sampling strategy to more frequently sample supervoxels containing less-frequent classes and faraway objects. 
On these supervoxels, we propose inter-point and inter-voxel affinity distillation, where the similarity information between points and voxels can help the student model better capture the structural information of the surrounding environment. We conduct extensive experiments on two popular LiDAR segmentation benchmarks, \ie, nuScenes~\cite{caesar2020nuscenes} and SemanticKITTI~\cite{behley2019semantickitti}. On both benchmarks, our \algorithmname consistently outperforms previous distillation approaches by a large margin on three representative backbones, \ie, Cylinder3D~\cite{zhu2021cylindrical,zhu2021cylindrical-tpami}, SPVNAS~\cite{tang2020searching} and MinkowskiNet~\cite{choy20194d}. Notably, on the challenging nuScenes and SemanticKITTI datasets, our method can achieve roughly \textbf{75\%} MACs reduction and \textbf{2}$\times$ speedup on the competitive Cylinder3D model and rank \textbf{1}st on the SemanticKITTI leaderboard among all published algorithms\footnote{https://competitions.codalab.org/competitions/20331\#results (single-scan competition) till 2021-11-18 04:00 Pacific Time, and our method is termed Point-Voxel-KD. Our method (PV-KD) ranks 3rd on the multi-scan challenge till 2021-12-1 00:00 Pacific Time.}. Our code is available at \url{https://github.com/cardwing/Codes-for-PVKD}. %Code and models will be released.
\end{abstract}

%////////////////////////////////
\section{Introduction}
\label{sec:introduction}
%////////////////////////////////
% !TEX root = ./main.tex

%\blfootnote{$\dagger$: Corresponding author.}

LiDAR semantic segmentation plays a vital role in the perception of autonomous driving as it provides per-point semantic information of the surrounding environment. With the advent of deep learning, plenty of LiDAR segmentation models have been proposed~\cite{zhu2021cylindrical,hu2020randla,tang2020searching,thomas2019kpconv} and have dominated the leaderboard of many benchmarks~\cite{behley2019semantickitti,caesar2020nuscenes}. However, the impressive performance comes at the expense of heavy computation and storage, which impedes them from being deployed in resource-constrained devices. 

%The latency of SPVNAS and Cylinder3D is 259 ms and 170 ms, respectively, which is far from real-time performance. 

To enable the deployment of these powerful LiDAR segmentation models on autonomous vehicles, knowledge distillation~\cite{hinton2015distilling} is a prevailing technique to transfer the dark knowledge from the overparameterized teacher model to the slim student network to achieve model compression. However, directly applying previous distillation algorithms~\cite{hinton2015distilling,liu2019structured,shu2020channel,wang2020intra,he2019knowledge} to LiDAR semantic segmentation brings marginal gains due to the intrinsic difficulty of point cloud, \ie, sparsity, randomness and varying density.

To address the aforementioned challenges, we propose the \textit{Point-to-Voxel Knowledge Distillation} (\algorithmname). As the name implies, we propose to distil the knowledge from both point-level and voxel-level. Specifically, to combat against the sparse supervision signals, we first propose to distil the pointwise and voxelwise probabilistic outputs from the teacher, respectively. The pointwise output contains fine-grained perceptual information while the voxelwise prediction embraces coarse but richer clues about the surrounding environment. 

To effectively distil the valuable structural knowledge from the unordered point sequences, we propose to exploit the point-level and voxel-level affinity knowledge. The affinity knowledge is obtained via measuring the pairwise semantic similarity of the point features and voxel features. However, straightforwardly mimicking the affinity knowledge of the whole point cloud is intractable since there are tens of thousands of points and the affinity matrix of these point features have over ten billion elements. Consequently, we put forward the supervoxel partition to divide the whole point cloud into a fixed number of supervoxels. At each distillation step, we only sample $K$ supervoxels and distil the affinity knowledge computed from point features and voxel features in these supervoxels, thus significantly enhancing the learning efficiency. 
Considering that uneven number-of-points distribution exists among different classes and objects at distinct distances, we further introduce a difficulty-aware sampling strategy to more frequently sample supervoxels that contain minority classes and faraway objects, emphasizing the learning on hard cases.
%Considering that severe quantity differences exist among different classes and objects at distinct distances, we further introduce a difficulty-aware sampling strategy to more frequently sample supervoxels that contain minority classes and faraway objects.

\begin{figure}[t]
 \centering
 \includegraphics[width=1.0\linewidth]{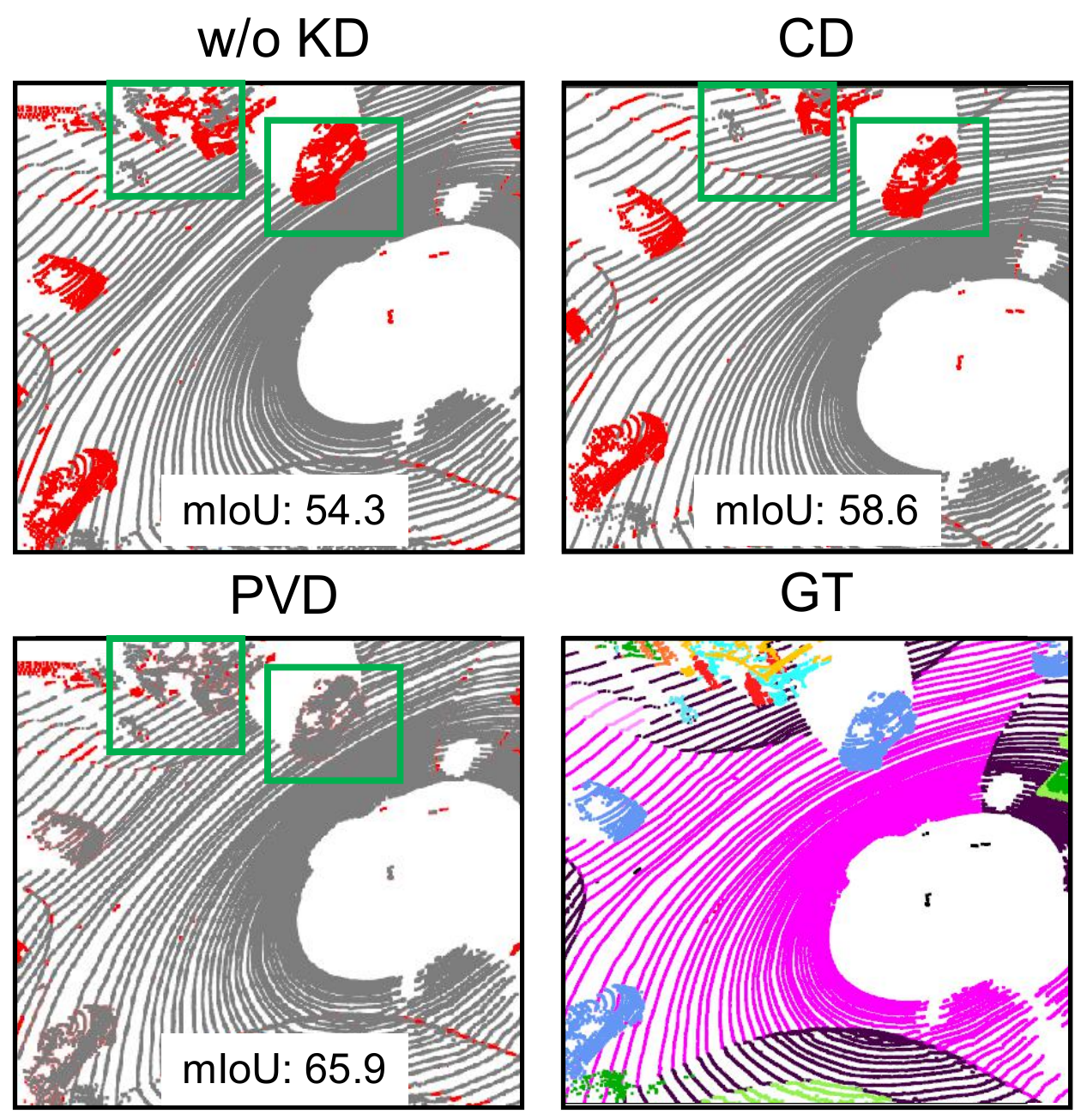}
 \vskip -0.2cm
 \caption{Comparison between the prediction of our method (\algorithmname) and the competitive channel distillation (CD) algorithm~\cite{shu2020channel} on SemanticKITTI validation set. Points that are mistakenly classified are painted red for better visualization. It is obvious that \algorithmname~can make the student model predict more accurately for those minority classes (person) and faraway objects (bicycles, highlighted with green rectangles) than the baseline distillation approach. Here, KD denotes knowledge distillation.}
 \centering
 \vskip -0.5cm
 \label{fig:motivation}
\end{figure}

We summarize our \textbf{contributions} as follow.
To our knowledge, we are the first to study how to apply knowledge distillation to LiDAR semantic segmentation for model compression.
To address the difficulty of distilling knowledge from point cloud, namely sparsity, randomness and varying density, we propose the point-to-voxel knowledge distillation. In addition, we put forward the supervoxel partition to make the affinity distillation tractable. A difficulty-aware sampling strategy is also employed to more frequently sample supervoxels that contain minority classes and distant objects, thus remarkably enhancing the distillation efficacy on these hard cases.
As can be seen from Fig.~\ref{fig:motivation}, our method produces much more accurate predictions than the baseline distillation approaches especially for those minority classes and faraway objects. 

We conduct extensive experiments on the nuScenes~\cite{caesar2020nuscenes} and SemanticKITTI~\cite{behley2019semantickitti} datasets and the results demonstrate that our algorithm consistently outperforms previous distillation approaches by a large margin on three contemporary models, \ie, Cylinder3D~\cite{zhu2021cylindrical,zhu2021cylindrical-tpami}, SPVNAS~\cite{tang2020searching} and MinkowskiNet~\cite{choy20194d}. Notably, on the nuScenes and SemanticKITTI benchmarks, \algorithmname~achieves approximately \textbf{75}\% MACs reduction and \textbf{2}$\times$ speedup on the top-performing Cylinder3D model with very minor performance degradation.

%The impressive model compression results indicate that there still exists large redundancy in contemporary LiDAR semantic segmentation models and we hope our approach can serve as a strong baseline to compress these cumbersome models.

%////////////////////////////////
\section{Related Work}
\label{sec:relatedwork}
%////////////////////////////////
% !TEX root = ./main.tex

\noindent \textbf{LiDAR semantic segmentation:} LiDAR semantic segmentation~\cite{zhu2021cylindrical,hu2020randla,cortinhal2020salsanext,qi2017pointnet,qi2017pointnet++,zhang2020polarnet,thomas2019kpconv,tang2020searching,Xu_2021_ICCV,choy20194d,xu2020squeezesegv3,wu2018squeezeseg,graham20183d,liu2019point} is crucial for the navigation of autonomous vehicles. PointNet~\cite{qi2017pointnet} is one of the pioneering work that uses Multi-Layer Perception (MLP) to process point cloud directly. Although effective in processing small-scale point cloud, PointNet and its variants~\cite{qi2017pointnet++} are extremely slow to handle large-scale outdoor point cloud. To cope with large-scale outdoor point cloud, Hu \etal~\cite{hu2020randla} exploit random sampling for point selection and a local feature aggregation module is designed to further preserve the key features. Xu \etal~\cite{Xu_2021_ICCV} put forward the range-point-voxel fusion network to make use of the advantages of different views. Zhu \etal~\cite{zhu2021cylindrical} propose the Cylinder3D method that adopts the cylindrical partition and asymmetrical convolution to better employ the valuable information in point cloud. Tang \etal~\cite{tang2020searching} leverage neural architecture search to automatically find the optimal structure for the task at hand. Although these models have shown impressive performance on various benchmarks, a common drawback is that these networks are too cumbersome to be deployed on resource-constrained devices. To enable the deployment of these powerful yet lumbersome LiDAR semantic segmentation models on real-world applications, we propose the point-to-voxel knowledge distillation to achieve model compression.

\noindent \textbf{Knowledge distillation:} Knowledge distillation (KD) stems from the seminal work of G. Hinton \etal~\cite{hinton2015distilling}. The primary objective of KD is to transfer the rich dark knowledge from a cumbersome teacher model to a compact student model to mitigate the performance gap between these two models. The majority of the KD approaches concentrate on image classification tasks and they take various forms of knowledge as the distillation targets, \eg, intermediate outputs~\cite{romero2015fitnets,hou2019learningto}, visual attention maps~\cite{zagoruyko2016paying,hou2019learning}, inter-layer similarity maps~\cite{yim2017gift}, sample-level similarity maps~\cite{park2019relational,tung2019similarity}, \etc. Recently, some researchers have adapted conventional KD techniques to distil knowledge for semantic segmentation tasks~\cite{liu2019structured,shu2020channel,wang2020intra,hou2020inter,he2019knowledge}. For instance, Liu \etal~\cite{liu2019structured} propose to distil three levels of knowledge simultaneously, namely the pixel-level knowledge, the pairwise similarity knowledge and the holistic knowledge. He \etal~\cite{he2019knowledge} make the student mimic the compressed knowledge as well as the affinity information of the teacher. Although previous distillation algorithms have shown excellent performance on 2D segmentation, straightforwardly deploying them on LiDAR segmentation tasks brings marginal gains owing to the inherent sparsity, randomness and varying density of point cloud. And to our best knowledge, we are the first to apply knowledge distillation to LiDAR semantic segmentation. The proposed \algorithmname~can effectively transfer both point-level and voxel-level knowledge to students and is suitable for distilling various LiDAR semantic segmentation models.
%\cavan{I suppose we have experiments to support this assertion}

%////////////////////////////////
\section{Methodology}
\label{sec:methodology}
%////////////////////////////////
% !TEX root = ./main.tex

Given an input point cloud $\mathbf{X} \in \mathbb{R}^{N \times 3}$, the objective of LiDAR semantic segmentation is to assign a class label $l \in \{0, 1, ..., C-1 \}$ to each point, where $N$ is the number of points and $C$ is the number of classes. Contemporary algorithms use CNNs for end-to-end prediction.

Considering that autonomous vehicles typically have limited computation and storage resources and call for real-time performance, efficient models are employed to fulfill the preceding requirement. Knowledge distillation~\cite{hinton2015distilling} is widely adopted to achieve model compression via conveying the rich dark knowledge from the large teacher model to the compact student network. However, previous distillation methods~\cite{liu2019structured,shu2020channel,wang2020intra,he2019knowledge} are tailored for 2D semantic segmentation tasks. Directly applying these algorithms to distill knowledge for 3D segmentation tasks produces unsatisfactory results owing to the intrinsic difficulty of point cloud, \ie, sparsity, randomness and varying density. To address the aforementioned challenges, we propose the Point-to-Voxel Knowledge Distillation (\algorithmname) to transfer the knowledge from both the point level and voxel level.

\begin{figure*}[t]
 \centering
 \includegraphics[width=1.0\linewidth]{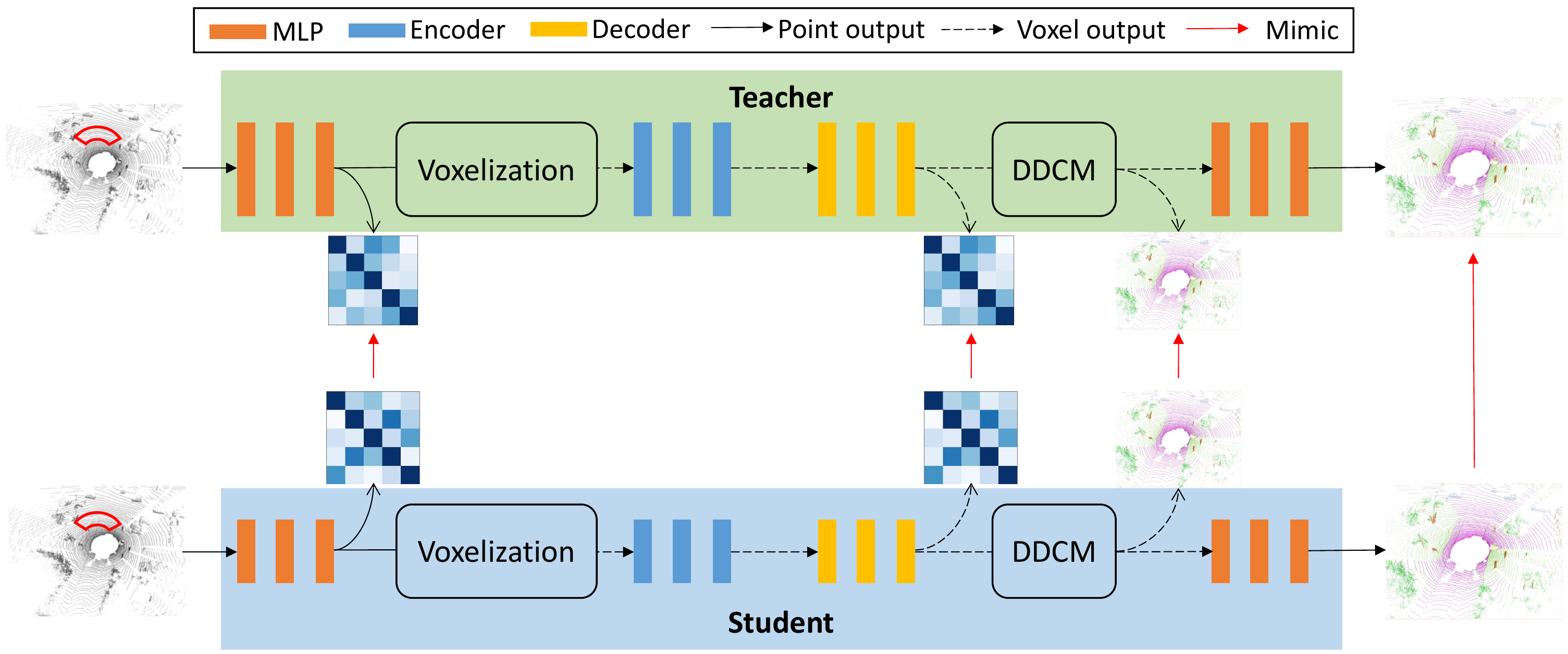}
 \vskip -0.2cm
 \caption{Framework overview. We take the Cylinder3D model~\cite{zhu2021cylindrical} as example. There are two networks in our framework, one is the teacher and the other is the student. The student model is obtained via pruning 50\% channels of each layer of the teacher. A teacher model is comprised of five parts, \ie, point feature extraction module, point-to-voxel transformation module (voxelization), an encoder-decoder model (asymmetric 3D convolution network), the Dimension-Decomposition Contextual Modeling (DDCM) module, and the point refinement module. Given the input point cloud, we first divide it into a fixed number of supervoxels and sample $K$ supervoxels according to the difficulty-aware sampling strategy ($K$=1 in this figure, denoted by the red sector). Then, the student is forced to mimic two levels of knowledge from the teacher. The first-level knowledge is the pointwise output and voxelwise output of the whole point cloud. The second-level knowledge is the inter-point affinity matrix and inter-voxel affinity matrix of the sampled supervoxels.}
 \centering
 \vskip -0.5cm
 \label{fig:framework_overview}
\end{figure*}

\subsection{Framework overview of Cylinder3D}

We first have a brief review of the Cylinder3D model~\cite{zhu2021cylindrical} and then introduce the point-to-voxel distillation algorithm based on it. Taking the point cloud as input, Cylinder3D first generates the corresponding feature for each point using a stack of MLPs and then reassigns the point features $F^{p} \in \mathbb{R}^{N \times C_{f}}$ based on the cylindrical partition where $C_{f}$ is the dimension of point features. Point features that belong to the same voxel are aggregated together through the maxpooling operation to obtain the voxel features $F^{v} \in \mathbb{R}^{M \times C_{f}}$ where $M$ is the number of non-empty voxels. Next, these voxel features are fed into the asymmetrical 3D convolution networks to produce the voxel-wise output $\mathbf{O}^{v} \in \mathbb{R}^{R \times A \times H \times C}$. A point-wise refinement module is further employed to produce the refined point-wise prediction $\mathbf{O}^{p} \in \mathbb{R}^{N \times C}$. Here, $N$, $C$, $R$, $A$ and $H$ denote the number of points, number of classes, radius, angle and height, respectively. Eventually, we will use the argmax operation to process the pointwise prediction to obtain the classification result of each point.

%Although Cylinder3D~\cite{zhu2021cylindrical} has achieved impressive performance in various LiDAR segmentation benchmarks, it requires extensive storage and computation resources, which impedes it from being deployed in resource-constrained devices. Knowledge distillation~\cite{hinton2015distilling} is a prevailing approach to achieve model compression via transferring the dark knowledge from the cumbersome teacher model to the compact student network. However, previous distillation methods are mainly designed for 2D segmentation tasks. Straightforwardly applying these approaches to LiDAR segmentation yields inferior results due to the intrinsic difficulty of point cloud, namely sparsity, randomness and varying density.

\subsection{Point-to-Voxel Output Distillation}
% Suppose the pointwise output and voxelwise output are $\mathbf{O}^{p} \in \mathbb{R}^{N \times C}$ and $\mathbf{O}^{v} \in \mathbb{R}^{R \times A \times H \times C}$, respectively.
The primary difference between 2D and 3D semantic segmentation lies in the input. Compared with images, point cloud is sparse and it is difficult to train the efficient student model using the sparse supervision signal. Previous distillation approaches~\cite{hinton2015distilling,liu2019structured} typically resort to distilling the ultimate output of the teacher network, \ie, the pointwise output of the teacher network for LiDAR semantic segmentation. Although the pointwise output contains fine-grained perceptual information of the environment, such knowledge is inefficient to learn as there are hundreds of thousands of points. To improve the learning efficiency, in addition to the pointwise output, we propose to distil the voxelwise output as the number of voxels is smaller and is easier to learn. The combination of both pointwise output distillation and voxelwise output distillation naturally form the coarse-to-fine learning process. The pointwise and voxelwise output distillation loss is given below:

\begin{equation}
\label{eqn:point_distillation_loss}
\begin{split}
\mathcal{L}_{out}^{p}(\mathbf{O}_{S}^{p}, \mathbf{O}_{T}^{p}) = \frac{1}{NC} \sum_{n=1}^{N} \sum_{c=1}^{C} \mathbf{KL}(\mathbf{O}_{S}^{p}(n, c) \| \mathbf{O}_{T}^{p}(n, c)),
\end{split}
\end{equation}

\begin{equation}
\label{eqn:voxel_distillation_loss}
\begin{split}
& \mathcal{L}_{out}^{v}(\mathbf{O}_{S}^{v}, \mathbf{O}_{T}^{v}) = \\
& \frac{1}{RAHC} \sum_{r=1}^{R} \sum_{a=1}^{A} \sum_{h=1}^{H} \sum_{c=1}^{C} \mathbf{KL}(\mathbf{O}_{S}^{v}(r, a, h, c) \| \mathbf{O}_{T}^{v}(r, a, h, c)),
\end{split}
\end{equation}
where $\mathbf{KL}(.)$ denotes the Kullback-Leibler divergence loss.

\noindent \textbf{Labels for the voxelwise output:} Since a voxel may contain points from different classes, how to assign the proper label to the voxel is also crucial to the performance. Following~\cite{zhu2021cylindrical}, we adopt the majority encoding strategy that uses the class label having the maximum number of points inside a voxel as the voxel label.
%The motivation for our sampling strategy is that supervoxels that contain infrequent classes or are far away from the origin should be more frequently sampled.

\subsection{Point-to-Voxel Affinity Distillation}

Distilling the knowledge of the pointwise and voxelwise outputs is insufficient as it merely considers the knowledge of each element and fails to capture the structural information of the surrounding environment. Such structural knowledge is vital to the LiDAR-based semantic segmentation model as the input points are unordered. A natural remedy is to adopt the relational knowledge distillation~\cite{liu2019structured} which calculates the pairwise similarity of all point features. However, there exist two shortcomings in this learning scheme: 1) since there are usually hundreds of thousands of points in an input point cloud, the similarity matrix which has over ten billion elements is computationally expensive to calculate and extremely difficult to learn. 2) There exist significant quantity differences between different classes and objects at different distances. The above-mentioned learning strategy ignores such difference and treats all classes and objects equally, thus making the distillation process sub-optimal.

\noindent \textbf{Supervoxel partition:} In order to more efficiently learn the relational knowledge, we divide the whole point cloud into several supervoxels whose size is $R_{s} \times A_{s} \times H_{s}$. Each supervoxel is comprised of a fixed number of voxels and the total number of supervoxels is $N_{s} = \lceil \frac{R}{R_{s}} \rceil \times \lceil \frac{A}{A_{s}} \rceil \times \lceil \frac{H}{H_{s}} \rceil$ where $\lceil . \rceil$ is the ceiling function. We will sample $K$ supervoxels to perform the affinity distillation.

\noindent \textbf{Difficulty-aware sampling:} To make supervoxels that contain less frequent classes and faraway objects more likely to be sampled, we present the difficulty-aware sampling strategy. The weight for choosing the $i$-th supervoxel is:
\begin{equation}
\label{eqn:supervoxel_prob}
\begin{split}
W_{i} = \frac{1}{f_{class}} \times \frac{d_{i}}{R} \times \frac{1}{N_{s}},
\end{split}
\end{equation}

\noindent where $f_{class}$ is the class frequency, $d_{i}$ is the distance of the outer arc of the $i$-th supervoxel to the origin in the X-Y plane. We treat the classes that have more than 1\% of all points in the whole dataset as majority classes and the remaining classes are considered as minority classes. We empirically set the class frequency of the supervoxel as: $f_{class} = 4\exp(-2N_{minor}) + 1$, where $N_{minor}$ is the number of voxels of minority classes in the supervoxel. If there is no voxel of minority class, the $f_{class}$ will be 5. And as the the number of voxels of minority classes increases, $f_{class}$ will be close to 1 quickly. Then, we normalize the weight and obtain the probability of the $i$-th supervoxel being sampled is: $P_{i} = \frac{W_{i}}{\sum_{i=1}^{N_{s}}W_{i}}$.

\begin{figure}[t]
 \centering
 \includegraphics[width=1.0\linewidth]{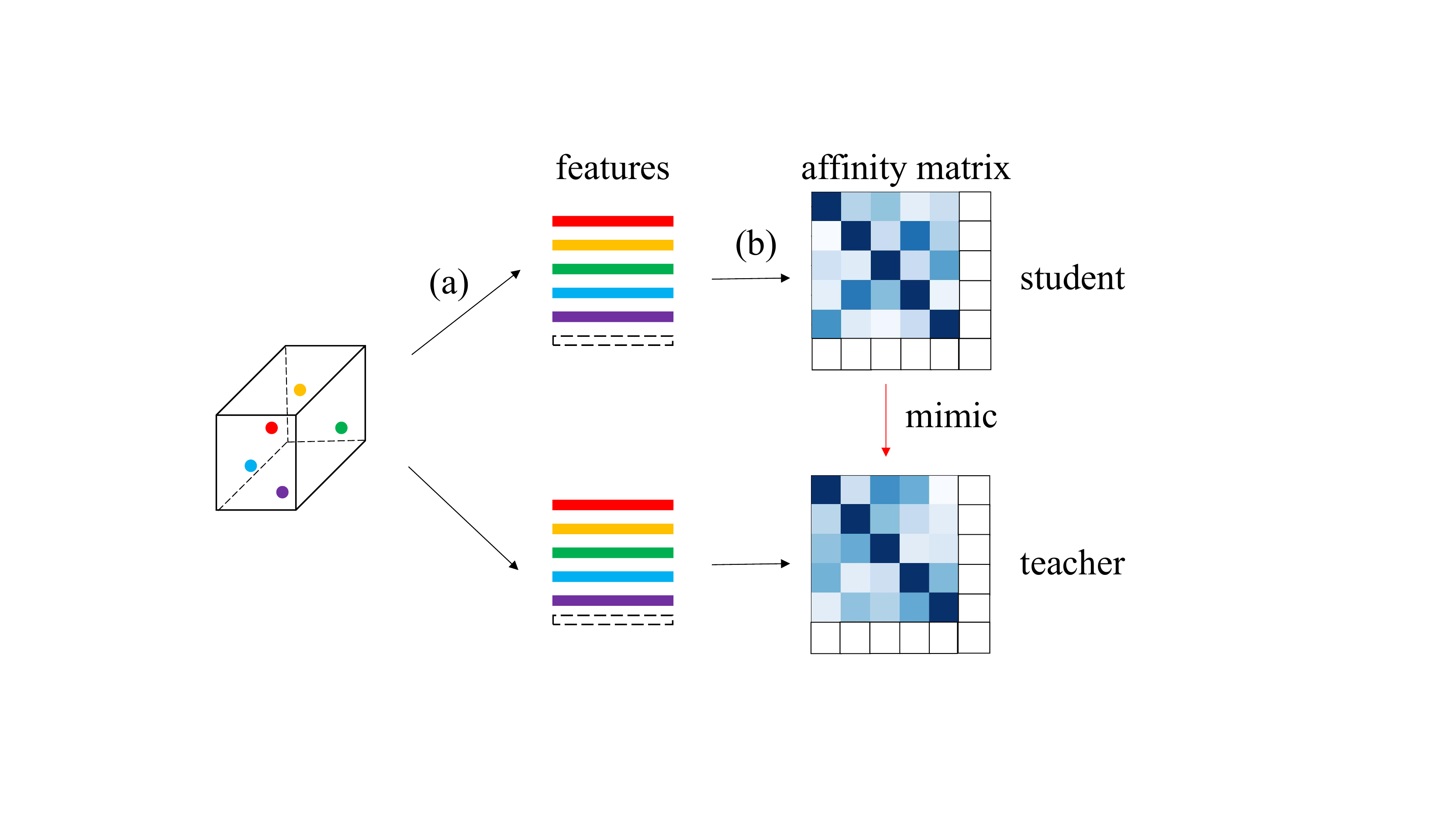}
 \vskip -0.3cm
 \caption{Computation of the inter-point affinity matrix within a supervoxel. Given the sampled supervoxel, we first (a) extract features from each point and then (b) obtain the affinity matrix via calculating the pairwise point features. Finally, the produced affinity matrix of the student is forced to mimic that of the teacher. Note that all-zeros features (the dashed rectangle) will be appended to current features if the number of point features is smaller than $N_{p}$.}
 \centering
 \vskip -0.6cm
 \label{fig:affinity_matrix}
\end{figure}

\noindent \textbf{Point/voxel features processing:} Note that for each point cloud, the number of input points is different and the density is varying, thus making the number of point features and voxel features variable in a supervoxel. As to the calculation of loss function, it is desirable to keep the number of features fixed. Hence, we set the number of retained point features and non-empty voxel features as $N_{p}$ and $N_{v}$, respectively. If the number of point features is larger than $N_{p}$, then we will retain $N_{p}$ point features by randomly discarding additional point features of majority class. If the number of point features is smaller than $N_{p}$, we will append all-zero features to the current features to obtain $N_{p}$ features, as is shown in Fig.~\ref{fig:affinity_matrix} (a). The voxel features are processed in a similar way.

Here we have $N_{p}$ point features $\hat{F}_{r}^{p} \in \mathbb{R}^{N_{p} \times C_{f}}$ and $N_{v}$ voxel features $\hat{F}_{r}^{v} \in \mathbb{R}^{N_{v} \times C_{f}}$ in the $r$-th supervoxel, respectively. Then, for each supervoxel, we calculate the inter-point affinity matrix according to the following equation:

\begin{equation}
\label{eqn:pt_aff_mat}
\begin{split}
\mathbf{C}^{p}(i, j, r) = \frac{\trans{\hat{F}_{r}^{p}(i)} \hat{F}_{r}^{p}(j)}{\| \hat{F}_{r}^{p}(i) \|_{2} \| \hat{F}_{r}^{p}(j) \|_{2}}, r \in \{1, ..., K\}
\end{split}
\end{equation}

The affinity score captures the similarity of each pair of point features and it can be taken as the high-level structural knowledge to be learned by the student. The inter-point affinity distillation loss is given as below:

\begin{equation}
\label{eqn:inter_point_kd_loss}
\begin{split}
& \mathcal{L}_{aff}^{p}(\mathbf{C}_{S}^{p}, \mathbf{C}_{T}^{p}) = \\
& \frac{1}{KN_{p}^{2}} \sum_{r=1}^{K} \sum_{i=1}^{N_{p}} \sum_{j=1}^{N_{p}} \| \mathbf{C}_{S}^{p}(i, j, r) - \mathbf{C}_{T}^{p}(i, j, r) \|_{2}^{2}.
\end{split}
\end{equation}

The inter-voxel affinity matrix is computed similarly. Eventually, we make the student mimic the generated affinity matrices of the teacher model. The inter-voxel affinity distillation loss is presented as follows:

\begin{equation}
\label{eqn:inter_voxel_kd_loss}
\begin{split}
& \mathcal{L}_{aff}^{v}(\mathbf{C}_{S}^{v}, \mathbf{C}_{T}^{v}) = \\ & \frac{1}{KN_{v}^{2}} \sum_{r=1}^{K} \sum_{i=1}^{N_{v}} \sum_{j=1}^{N_{v}} \| \mathbf{C}_{S}^{v}(i, j, r) - \mathbf{C}_{T}^{v}(i, j, r) \|_{2}^{2}.
\end{split}
\end{equation}

\begin{figure}[t]
 \centering
 \includegraphics[width=0.85\linewidth]{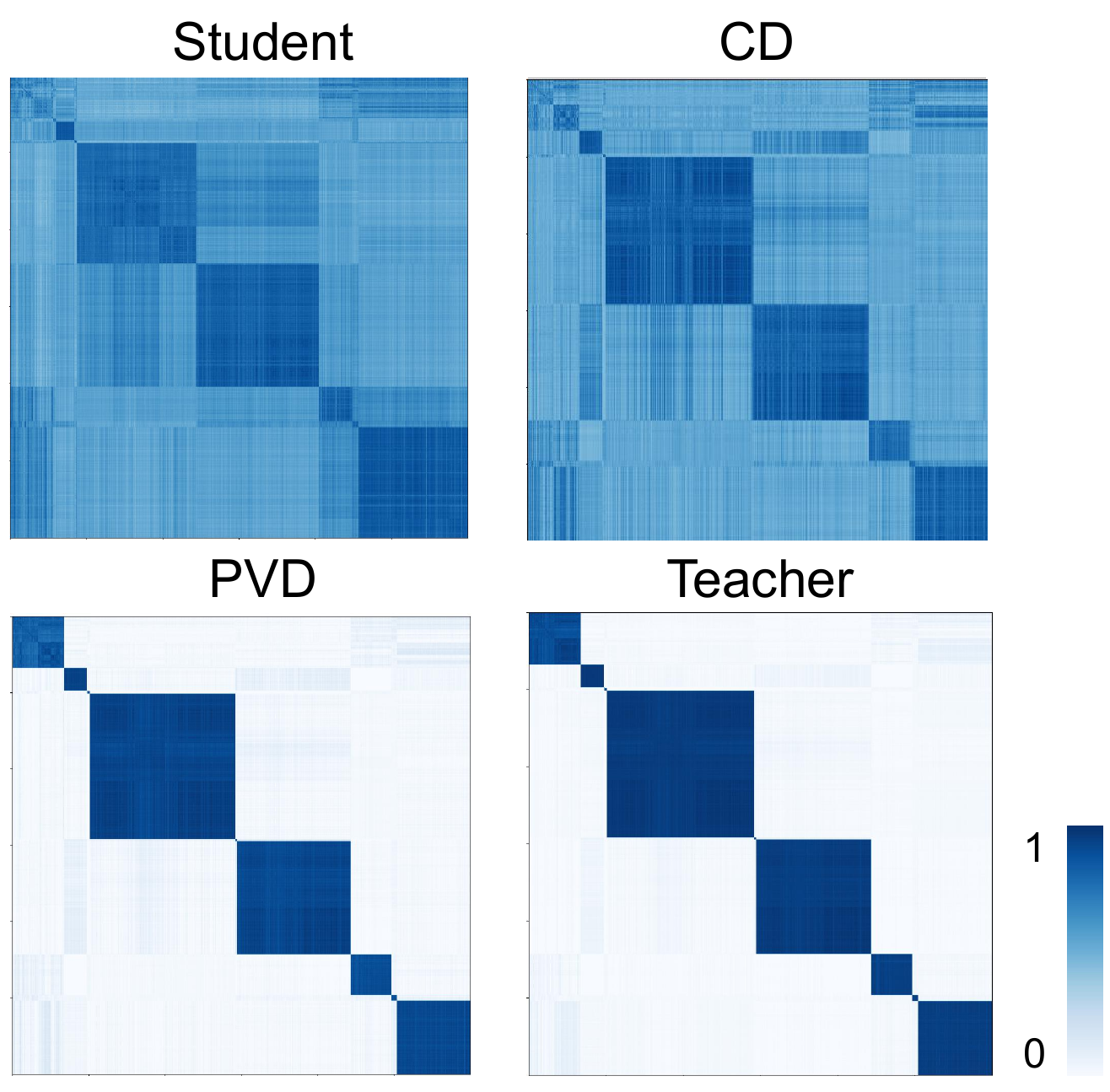}
 \vskip -0.2cm
 \caption{Comparison between inter-voxel affinity maps of different algorithms. The affinity value is obtained via normalizing the cosine similarity score to [0, 1].}
 \centering
 \vskip -0.6cm
 \label{fig:affinity_map_vis}
\end{figure}

\noindent \textbf{Visualization of the learned affinity maps:} From Fig.~\ref{fig:affinity_map_vis}, we can see that \algorithmname~causes a closer affinity map between student and teacher. With \algorithmname~, features that belong to the same class are pulled closer while those of different classes are pushed apart in the feature space, resulting in a much clear affinity map. Compared with the rival channel distillation approach~\cite{shu2020channel}, \algorithmname~can better transfer the structural knowledge from teacher to student, which strongly validates the superiority of \algorithmname~in distilling LiDAR segmentation models.

\subsection{Final objective}

Our final loss function is composed of seven terms, \ie, the weighted cross entropy loss for the pointwise output and voxelwise output, the lovasz-softmax loss~\cite{berman2018the}, the point-to-voxel output distillation loss and the point-to-voxel affinity distillation loss:

\begin{equation}
\label{eqn:loss_func}
\begin{split}
\mathcal{L} = & \mathcal{L}_{wce}^{p} + \mathcal{L}_{wce}^{v} + \mathcal{L}_{lovasz} + \alpha_{1} \mathcal{L}_{out}^{p}(\mathbf{O}_{S}^{p}, \mathbf{O}_{T}^{p}) \\ 
& + \alpha_{2} \mathcal{L}_{out}^{v}(\mathbf{O}_{S}^{v}, \mathbf{O}_{T}^{v}) + \beta_{1} \mathcal{L}_{aff}^{p}(\mathbf{C}_{S}^{p}, \mathbf{C}_{T}^{p}) \\
& + \beta_{2}\mathcal{L}_{aff}^{v}(\mathbf{C}_{S}^{v}, \mathbf{C}_{T}^{v}),
\end{split}
\end{equation}

\noindent where $\alpha_{1}$, $\alpha_{2}$, $\beta_{1}$ and $\beta_{2}$ are the loss coefficients to balance the effect of the distillation losses on the main task loss.

%////////////////////////////////
\section{Experiments}
\label{sec:experiments}
%////////////////////////////////
% !TEX root = ./main.tex

\noindent \textbf{Datasets.} Following the practice of Cylinder3D~\cite{zhu2021cylindrical}, we conduct experiments on two popular LiDAR semantic segmentation benchmarks, \ie, nuScenes~\cite{caesar2020nuscenes} and SemanticKITTI~\cite{behley2019semantickitti}. For nuScenes, it consists of 1000 driving scenes, in which 850 scenes are selected for training and validation, and the remaining 150 scenes are chosen for testing. 16 classes are utilized for LiDAR semantic segmentation after merging similar classes and eliminating infrequent classes. For SemanticKITTI, it is comprised of 22 point cloud sequences, where sequences 00 to 10, 08 and 11 to 21 are used for training, validation and testing, respectively. A total number of 19 classes are chosen for training and evaluation after merging classes with distinct moving status and discarding classes with very few points.

\noindent \textbf{Evaluation metrics.} Following~\cite{zhu2021cylindrical}, we adopt the intersection-over-union (IoU) of each class and mIoU of all classes as the evaluation metric. The calculation of IoU is: $IoU_{i} = \frac{TP_{i}}{TP_{i}+FP_{i}+FN_{i}}$, where $TP_{i}$, $FP_{i}$ and $FN_{i}$ represent the true positive, false positive and false negative of class $i$.

\noindent \textbf{Implementation details.} Following~\cite{zhu2021cylindrical}, we leverage Adam~\cite{kingma2015adam} as the optimizer and the initial learning rate is set as 2e-3. Batch size is set as 4 and the number of training epochs is 40. $\alpha_{1}$, $\alpha_{2}$, $\beta_{1}$ and $\beta_{2}$ are set as 0.1, 0.15, 0.15 and 0.25, respectively. We take the rival and open sourced Cylinder3D\footnote{https://github.com/xinge008/Cylinder3D}~\cite{zhu2021cylindrical} approach as the backbone since the top-performing RPVNet~\cite{Xu_2021_ICCV} and AF2S3Net~\cite{cheng20212} do not release their codes. Random flipping, rotation, scaling and transformation are taken as the data augmentation strategy. The size of the voxel output is 480$\times$360$\times$32, where the three dimensions denote the radius, angle and height, respectively. The size of the supervoxel is set as 120$\times$60$\times$8. $N_{v}$ and $N_{p}$ are set as 3000 and 6000, respectively. The number of sampled supervoxels $K$ is set as 4. The inter-point affinity distillation is performed on the output of the point feature extraction module and the inter-voxel affinity distillation is conducted on the output of the encoder-decoder backbone. For nuScenes, Cylinder3D\_0.5$\times$ is produced from the original Cylinder3D model by pruning 50\% channels for each layer of the whole network. For SemanticKITTI, Cylinder3D\_0.5$\times$ is obtained by merely pruning 50\% channels for each layer of the asymmetrical 3D convolution network and we keep the point feature extraction module unchanged as it is vital to extract rich information from the input point cloud. %Note that although we mainly perform experiments on the Cylinder3D model, our method is applicable to all voxel-based LiDAR semantic segmentation methods.
We also apply our method to compress SPVNAS\footnote{https://github.com/mit-han-lab/spvnas}~\cite{tang2020searching} and MinkowskiNet~\cite{choy20194d} to verify the scalability of our algorithm. More details are provided in the supplementary material. 

\noindent \textbf{Baseline distillation algorithms.} In addition to the state-of-the-art methods in each benchmark, we also compare our method with classical KD methods and contemporary distillation approaches tailored for 2D semantic segmentation, including vanilla KD~\cite{hinton2015distilling}, SKD~\cite{liu2019structured}, CD~\cite{shu2020channel}, IFV~\cite{wang2020intra} and KA~\cite{he2019knowledge}. Here, SKD takes the output probability maps and pairwise similarity maps as mimicking targets. We remove the original holistic distillation loss for SKD as incorporating GANs into current framework will cause severe training instability; CD utilizes the intermediate feature maps and score maps as knowledge; IFV transfers the intra-class feature variation from the teacher to the student; KA makes the student distil the compressed knowledge and the affinity information of the whole output from the teacher.

\subsection{Results}

%\noindent \textbf{Quantitative results:}

\noindent \textbf{Comparison with state-of-the-art LiDAR segmentation models:} We compare our model with contemporary LiDAR semantic segmentation models, \eg, KPConv~\cite{thomas2019kpconv}, TORNADONet~\cite{gerdzhev2021tornado} and SPVNAS~\cite{tang2020searching}. From Table~\ref{semantickitti}, we can see that Cylinder3D\_0.5$\times$+\algorithmname~(the penultimate row) achieves comparable performance with the original Cylinder3D model on the SemanticKITTI test set. Compared to KPConv and SPVNAS, our Cylinder3D\_0.5$\times$+\algorithmname~not only achieves better performance, \eg, \textbf{3.8\%} higher than SPVNAS in mIoU, but also has much lower latency than the SPVNAS method (259 ms v.s. 76 ms). Specifically, on minority classes such as bicycle, motocycle and bicyclist, the IoU of Cylinder3D\_0.5$\times$+\algorithmname~is at least \textbf{10.5\%} higher than the SPVNAS method. And with some engineering tricks like finetuning and flip \& rotation test ensemble, our Cylinder3D\_0.5$\times$+\algorithmname~(the last row) can obtain \textbf{71.2} mIoU, which is \textbf{2.3} mIoU higher than the original Cylinder3D model. Impressive performance is also observed in nuScenes validation set. Our Cylinder3D\_0.5$\times$+\algorithmname~exhibits similar performance with the original Cylinder3D network in terms of the overall mIoU and the IoU on each class.

\noindent \textbf{Comparison with previous distillation methods:} From Table~\ref{semantickitti} and~\ref{nuscenes}, we can see that \algorithmname~significantly outperforms baseline distillation algorithms in both benchmarks. The performance gap between \algorithmname~and the most competitive KD method is larger than 1.8. For instance, on SemanticKITTI test set, our \algorithmname~is \textbf{2.8} mIoU higher than the CD method. And on both majority classes and minority classes, our \algorithmname~significantly outperforms traditional distillation algorithms. For instance, on nuScenes dataset, \algorithmname~is at least \textbf{2} mIoU higher than SKD in classes such as bicycle, bus, car, trailer and sidewalk. The aforementioned results strongly demonstrate the effectiveness of \algorithmname~in transferring knowledge for teacher-student learning.

\begin{table*}[t]
\caption{Quantitative results of our proposed method and state-of-the-art LiDAR semantic segmentation methods as well as previous distillation approaches on SemanticKITTI test set. Cylinder3D\_0.5$\times$ is abbreviated as C3D\_0.5$\times$ to save space. * means that finetuning and flip \& rotation test ensemble are applied. All results can be found in the online leaderboard.}
\vskip -0.2cm
\label{semantickitti}
\centering
\begin{adjustbox}{width=\textwidth}
\begin{tabular}{c|c|c|c|c|c|c|c|c|c|c|c|c|c|c|c|c|c|c|c|c|c}
\hline
\textbf{Methods} & \textbf{mIoU} & \rotatebox{90}{Latency (ms)} & \rotatebox{90}{car} &  \rotatebox{90}{bicycle} & \rotatebox{90}{motorcycle} & \rotatebox{90}{truck} & \rotatebox{90}{other-vehicle} & \rotatebox{90}{person} & \rotatebox{90}{bicyclist} & \rotatebox{90}{motorcyclist} & \rotatebox{90}{road} & \rotatebox{90}{parking} & \rotatebox{90}{sidewalk} & \rotatebox{90}{other-ground} &
\rotatebox{90}{building} & \rotatebox{90}{fence} & \rotatebox{90}{vegetation} & \rotatebox{90}{trunk} & \rotatebox{90}{terrain} & \rotatebox{90}{pole} & \rotatebox{90}{traffic} \\
\hline
\hline
%RandLA-Net~\cite{hu2020randla} & 50.3 & 800 & 94.0 & 19.8 & 21.4 & {42.7} & 38.7 & 47.5 & 48.8 & 4.6  & 90.4 & 56.9 & 67.9 & 15.5 & 81.1 & 49.7 & 78.3 & 60.3 & 59.0 & 44.2 & 38.1 \\
%\hline
%RangeNet++~\cite{milioto2019rangenet++} & 52.2 & -- & 91.4 & 25.7 & 34.4 & 25.7 & 23.0 & 38.3 &  38.8 & 4.8 & {91.8} & {65.0} & 75.2 & 27.8 & 87.4 & 58.6 & 80.5 & 55.1 & 64.6 & 47.9 & 55.9 \\
%\hline
%PolarNet~\cite{zhang2020polarnet} & 54.3 & -- & 93.8 & 40.3 & 30.1 & 22.9 & 28.5 & 43.2 & 40.2 & 5.6 & 90.8 & 61.7 & 74.4 & 21.7 & {90.0} & 61.3 & 84.0 & 65.5 & 67.8 & 51.8 & 57.5  \\
%\hline
%SqueezeSegv3~\cite{xu2020squeezesegv3} & 55.9 & -- & 92.5 & 38.7 & 36.5 & 29.6 & 33.0 & 45.6 & 46.2 & {20.1} & 91.7 & 63.4 & 74.8 & 26.4 & 89.0 & 59.4 & 82.0 & 58.7 & 65.4 & 49.6 & 58.9  \\
%\hline
Salsanext~\cite{cortinhal2020salsanext} & 59.5 & -- & 91.9 & 48.3 & 38.6 & 38.9 & 31.9 & 60.2 & 59.0 & 19.4 & 91.7 & 63.7 & 75.8 & 29.1 & 90.2 & 64.2 & 81.8 & 63.6 & 66.5 & 54.3 & 62.1 \\
\hline
KPConv~\cite{thomas2019kpconv} & 58.8 & 263 & 96.0 & 32.0 & 42.5 & 33.4&44.3&61.5 & 61.6 & 11.8 & 88.8 & 61.3&  72.7&31.6& \bf{95.0} & 64.2 & 84.8 & 69.2 & 69.1 & 56.4 & 47.4 \\
\hline
FusionNet~\cite{zhang2020deep} * & 61.3 & -- & 95.3 & 47.5 & 37.7 & 41.8 & 34.5 & 59.5 & 56.8 & 11.9 & 91.8 & 68.8 & 77.1 & 30.8 & 92.5 & \bf{69.4} & 84.5 & 69.8 & 68.5&60.4 & 66.5 \\ 
\hline
KPRNet~\cite{kochanov2020kprnet} * & 63.1 & -- & 95.5&54.1& 47.9&23.6 & 42.6&65.9 & 65.0 & 16.5 & \bf{93.2} & \bf{73.9} & \bf{80.6} & 30.2 & 91.7 & {68.4} & 85.7 & 69.8 & 71.2 & 58.7 & 64.1 \\
\hline
TORNADONet~\cite{gerdzhev2021tornado} * & 63.1 & --&94.2& 55.7& 48.1& 40.0& 38.2& 63.6& 60.1& 34.9& 89.7& 66.3& 74.5& 28.7& 91.3& 65.6& 85.6& 67.0& 71.5 & 58.0 & {65.9} \\
\hline
SPVNAS~\cite{tang2020searching} * & 66.4 & 259 & \bf{97.3} & 51.5 & 50.8 & 59.8 & 58.8 & 65.7 & 65.2 & 43.7 & 90.2 & 67.6 & 75.2 & 16.9 & 91.3 & 65.9 & 86.1 & 73.4 & 71.0 & 64.2 & 66.9 \\
\hline
%Cylinder3D~\cite{zhu2021cylindrical} & \bf{67.8} & 97.1 & \bf{67.6} & \bf{64.0} & 59.0 & 58.6 & \bf{73.9} & \bf{67.9} & 36.0 & {91.4} & {65.1} & {75.5} & \bf{32.3} & {91.0} & {66.5} & {85.4} & 71.8 & {68.5} & 62.6 & {65.6}  \\
Cylinder3D~\cite{zhu2021cylindrical} * & 68.9 & 170 & 97.1 & 67.6 & 63.8 & 50.8 & 58.5 & 73.7 & 69.2 & 48.0 & 92.2 & 65.0 & 77.0 & 32.3 & 90.7 & 66.5 & 85.6 & 72.5 & 69.8 & 62.4 & 66.2 \\
\hline
C3D\_0.5$\times$ & 65.3 & \multirow{8}*{\textbf{76}} & 93.4 & 62.3 & 59.2 & 48.3 & 56.4 & 72.3 & 66.3 & 21.0 & 91.2 & 61.3 & 75.3 & 30.4 & 89.8 & 65.4 & 84.2 & 71.4 & 67.3 & 60.2 & 64.2 \\
\cline{1-2}\cline{4-22}
C3D\_0.5$\times$ + KD & 65.6 & ~ & 93.8 & 62.5 & 59.4 & 48.6 & 55.3 & 72.9 & 66.5 & 21.9 & 91.8 & 61.3 & 75.7 & 30.5 & 90.4 & 65.5 & 84.3 & 71.7 & 67.6 & 60.3 & 64.8 \\
\cline{1-2}\cline{4-22}
C3D\_0.5$\times$ + CD & 66.1 & ~ & 94.5 & 62.7 & 59.8 & 49.3 & 57.2 & 72.1 & 67.1 & 22.7 & 92.1 & 61.4 & 74.9 & 30.8 & 90.9 & 67.3 & 84.6 & 72.2 & 68.3 & 61.1 & 65.1 \\
\cline{1-2}\cline{4-22}
C3D\_0.5$\times$ + IFV & 65.5 & ~ & 93.7 & 62.4 & 59.1 & 48.7 & 56.7 & 72.4 & 66.6 & 21.4 & 91.5 & 62.0 & 75.6 & 30.4 & 90.3 & 66.2 & 84.7 & 71.5 & 67.5 & 60.6 & 64.3 \\
\cline{1-2}\cline{4-22}
C3D\_0.5$\times$ + SKD & 65.8 & ~ & 93.6 & 62.7 & 59.6 & 48.5 & 57.4 & 72.8 & 66.7 & 24.3 & 91.6 & 61.4 & 75.9 & 30.8 & 90.1 & 65.6 & 84.8 & 71.7 & 67.5 & 60.7 & 65.2 \\
\cline{1-2}\cline{4-22}
C3D\_0.5$\times$ + KA & 65.5 & ~ & 93.4 & 62.6 & 58.9 & 48.5 & 56.5 & 72.7 & 66.5 & 20.7 & 91.6 & 61.5 & 75.3 & 30.1 & 89.9 & 65.6 & 84.4 & 71.3 & 67.8 & 60.3 & 65.8 \\
\cline{1-2}\cline{4-22}
\bf{C3D\_0.5$\times$ + \algorithmname} & 68.9 & ~ & 96.7 & 66.4 & 61.0 & \bf{60.0} & 59.3 & 73.2 & 72.1 & 25.0 & 91.4 & 66.5 & 76.2 & 37.1 & 93.0 & 70.5 & 85.9 & 72.7 & 69.8 & 64.1 & \bf{67.8} \\
\cline{1-2}\cline{4-22}
%\bf{C3D\_0.5$\times$ + \algorithmname~*} & \bf{70.8} & ~ & 96.8 & 67.1 & \bf{70.0} & 55.3 & \bf{60.1} & \bf{74.4} & \bf{73.6} & \bf{49.4} & 91.4 & 69.6 & 76.5 & \bf{39.7} & 92.4 & 69.2 & \bf{86.3} & \bf{73.6} & 71.1 & 64.1 & 65.0 \\
\bf{C3D\_0.5$\times$ + \algorithmname~*} & \bf{71.2} & ~ & 97.0 & \bf{67.9} & \bf{69.3} & 53.5 & \bf{60.2} & \bf{75.1} & \bf{73.5} & \bf{50.5} & 91.8 & 70.9 & 77.5 & \bf{41.0} & 92.4 & \bf{69.4} & \bf{86.5} & \bf{73.8} & \bf{71.9} & \bf{64.9} & 65.8 \\
\hline
\end{tabular}
\end{adjustbox}
\vspace{-2ex}
\end{table*}

%Our method is named "Point-Voxel-KD" on the SemanticKITTI leaderboard\footnote{https://competitions.codalab.org/competitions/20331#results}

\begin{table*}[t]
\caption{Quantitative results of our proposed method and state-of-the-art LiDAR semantic segmentation methods as well as previous distillation approaches on nuScenes validation set.}
\vskip -0.2cm
\label{nuscenes}
\centering
\begin{adjustbox}{width=\textwidth}
\begin{tabular}{c|c|c|c|c|c|c|c|c|c|c|c|c|c|c|c|c|c}
\hline
\textbf{Methods} & \textbf{mIoU} & \rotatebox{90}{barrier} &  \rotatebox{90}{bicycle} & \rotatebox{90}{bus} & \rotatebox{90}{car} & \rotatebox{90}{construction} & \rotatebox{90}{motorcycle} & \rotatebox{90}{pedestrian} & \rotatebox{90}{traffic-cone} & \rotatebox{90}{trailer} & \rotatebox{90}{truck} & \rotatebox{90}{driveable} & \rotatebox{90}{other} &
\rotatebox{90}{sidewalk} & \rotatebox{90}{terrain} & \rotatebox{90}{manmade} & \rotatebox{90}{vegetation} \\
\hline
\hline
RangeNet++~\cite{milioto2019rangenet++} & 65.5 & 66.0 & 21.3 & 77.2 & 80.9 & 30.2 & 66.8 & 69.6 &  52.1 & 54.2 & {72.3} & {94.1} & 66.6 & 63.5 & 70.1 & 83.1 & 79.8 \\
\hline
PolarNet~\cite{zhang2020polarnet} & 71.0 & 74.7 & 28.2 & 85.3 & 90.9 & 35.1 & 77.5 & 71.3 & 58.8 & 57.4 & 76.1 & 96.5 & 71.1 & 74.7 & {74.0} & 87.3 & 85.7  \\
\hline
Salsanext~\cite{cortinhal2020salsanext} & 72.2 & 74.8 & 34.1 & 85.9 & 88.4 & 42.2 & 72.4 & 72.2 & 63.1 & 61.3 & 76.5 & 96.0 & 70.8 & 71.2 & 71.5 & 86.7 & 84.4 \\
\hline
\hline
Cylinder3D~\cite{zhu2021cylindrical} & \bf{76.1} & \bf{76.4} & \bf{40.3} & \bf{91.2} & 93.8 & \textbf{51.3} & \bf{78.0} & \bf{78.9} & \bf{64.9} & \bf{62.1} & \bf{84.4} & \bf{96.8} & \bf{71.6} & \bf{76.4} & 75.4 & \bf{90.5} & \bf{87.4}  \\
\hline
C3D\_0.5$\times$ & 73.6 & 74.6 & 36.2 & 88.2 & 87.3 & 47.9 & 76.4 & 77.0 & 63.4 & 58.8 & 82.3 & 95.1 & 70.0 & 73.5 & 73.6 & 88.7 & 85.2 \\
\hline
C3D\_0.5$\times$ + KD & 73.9 & 75.2 & 35.4 & 88.3 & 88.2 & 47.6 & 76.8 & 77.2 & 63.6 & 57.3 & 83.1 & 95.7 & 70.1 & 75.2 & 73.1 & 89.2 & 85.3 \\
\hline
C3D\_0.5$\times$ + CD & 74.1 & 75.4 & 36.1 & 88.4 & 89.3 & 46.9 & 76.1 & 77.6 & 62.9 & 58.0 & 84.3 & 96.0 & 70.3 & 74.8 & 74.6 & 90.1 & 85.6 \\
\hline
C3D\_0.5$\times$ + IFV & 73.8 & 74.7 & 36.6 & 88.3 & 88.6 & 47.2 & 76.7 & 77.1 & 63.1 & 58.2 & 83.5 & 95.1 & 70.2 & 73.4 & 73.8 & 88.9 & 84.3 \\
\hline
C3D\_0.5$\times$ + SKD & 74.2 & 74.9 & 37.3 & 87.6 & 89.1 & 47.5 & 76.2 & 77.4 & 63.2 & 59.3 & 83.4 & 95.9 & 70.4 & 73.9 & 74.3 & 90.3 & 87.1 \\
\hline
C3D\_0.5$\times$ + KA & 73.9 & 74.2 & 36.3 & 88.5 & 87.6 & 47.1 & 76.9 & 78.3 & 63.5 & 57.6 & 83.4 & 94.9 & 70.3 & 73.8 & 73.2 & 88.4 & 86.3 \\
\hline
\bf{C3D\_0.5$\times$ + \algorithmname~} & 76.0 & 76.2 & 40.0 & 90.2 & \bf{94.0} & 50.9 & 77.4 & 78.8 & 64.7 & 62.0 & 84.1 & 96.6 & 71.4 & 76.4 & \bf{76.3} & 90.3 & 86.9 \\
\hline
\end{tabular}
\end{adjustbox}
\vspace{-3ex}
\end{table*}

\begin{comment}
C3D\_0.5$\times$ \\
\hline
\end{comment}

\begin{table}[!t]
\caption{Performance of different algorithms on compressing SPVNAS and MinkowskiNet on SemanticKITTI validation set.}
\vskip -0.3cm
\label{spvnas_table}
\centering
\small{
\begin{tabular}{c|c|c}
\hline
Algorithm & mIoU & MACs (G) \\
\hline
\hline
SPVNAS~\cite{tang2020searching} & \bf{63.8} & 118.6 \\
\hline
SPVNAS$\_$0.5$\times$ & 60.4 & \multirow{4}*{29.7}  \\
\cline{1-2}
SPVNAS$\_$0.5$\times$ + CD & 60.9 & ~ \\
\cline{1-2}
SPVNAS$\_$0.5$\times$ + SKD & 61.2 & ~ \\
\cline{1-2}
\textbf{SPVNAS$\_$0.5$\times$ + \algorithmname~} & \textbf{63.8} & ~ \\
\hline \hline
MinkowskiNet~\cite{choy20194d} & 61.9 & 114.0 \\
\hline
MinkowskiNet$\_$0.5$\times$ & 58.9 & \multirow{4}*{28.5}  \\
\cline{1-2}
MinkowskiNet$\_$0.5$\times$ + CD & 59.6 & ~ \\
\cline{1-2}
MinkowskiNet$\_$0.5$\times$ + SKD & 59.4 & ~ \\
\cline{1-2}
\textbf{MinkowskiNet$\_$0.5$\times$ + \algorithmname~} & 61.8 & ~ \\
\hline
\end{tabular}
}
\vspace{-4ex}
\end{table}

\noindent \textbf{Generalization to more architectures:} To verify the generalization of our method, we also apply \algorithmname~to compress SPVNAS~\cite{tang2020searching} and MinkowskiNet~\cite{choy20194d}. Since SPVNAS does not provide the training code for the NAS-based architecture, we conduct experiments on its manually designed architecture. As can be seen from Table~\ref{spvnas_table}, our \algorithmname~still brings more gains to the student model than baseline distillation algorithms. For instance, our \algorithmname~outperforms the SKD algorithm by \textbf{2.6} mIoU in terms of mIoU on the SPVNAS backbone. It is noteworthy that \algorithmname~can safely achieve 75\% MACs reduction without causing severe performance drop. The above results strongly demonstrate the good scalability of our method.

\begin{figure*}[t]
 \centering
 \includegraphics[width=1.0\linewidth]{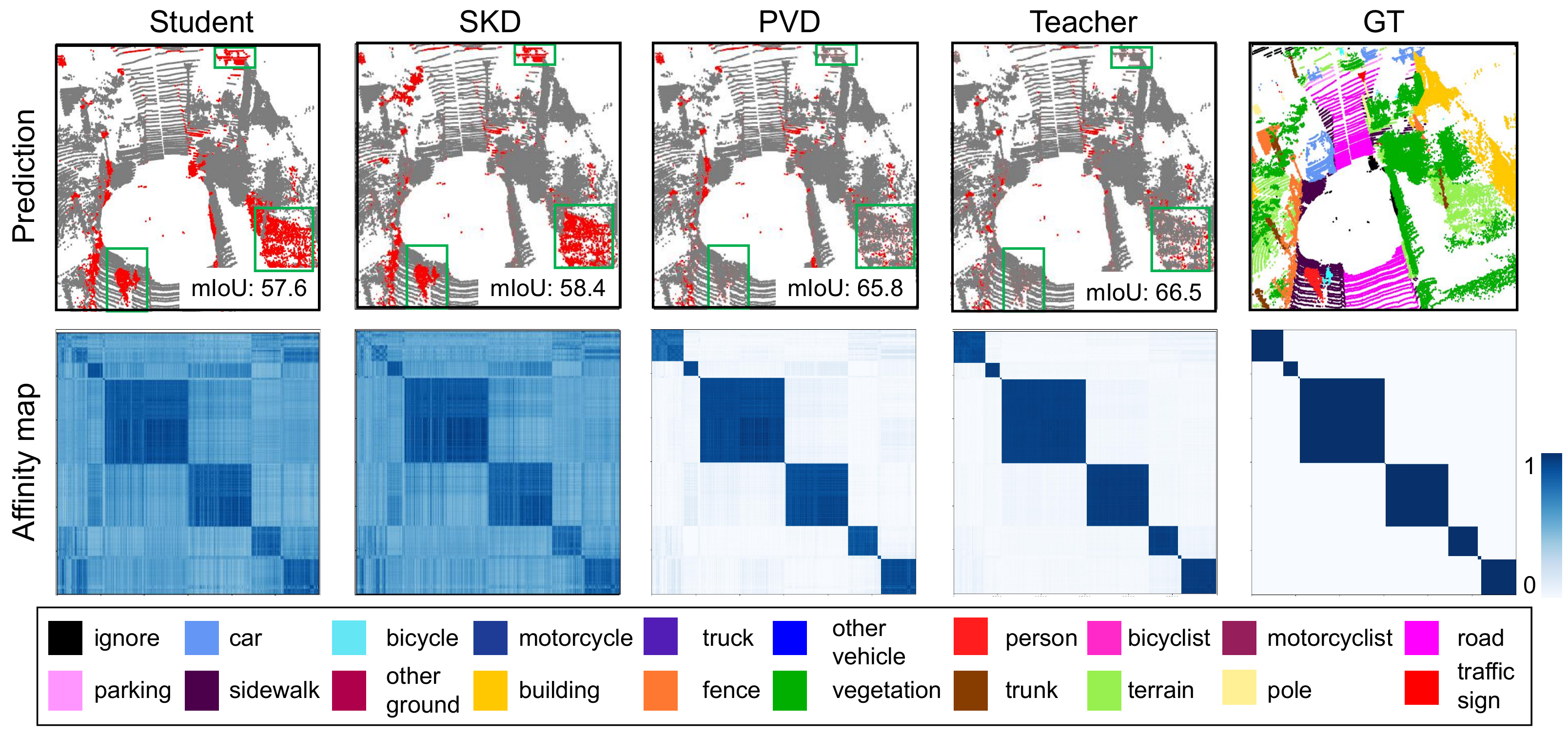}
 \vskip -0.3cm
 \caption{Visual comparison of different methods on the SemanticKITTI validation set. Here, the ground-truth for the inter-voxel affinity map is the ideal map where the intra-class similarity score is 1 and inter-class similarity score is 0.}
 \centering
 \vskip -0.6cm
 \label{fig:visual_compare}
\end{figure*}

\noindent \textbf{Qualitative results:} As can be seen from Fig.~\ref{fig:visual_compare}, compared with the SKD approach, our \algorithmname~greatly improves the prediction of the student model. The prediction errors of \algorithmname~on those minority classes, \eg, person and bicycle, are significantly smaller than those of SKD. Besides, on objects that are faraway from the origin, \eg, the car highlighted by the green rectangle, \algorithmname~also yields more accurate predictions than SKD. And \algorithmname~has lower inter-class similarity and higher intra-class similarity, which explicitly showcases the efficacy of \algorithmname~in distilling structural knowledge from the teacher model.     %In addition to quantitative comparison, we also provide visual comparison of the \algorithmname~with the SKD method.

\noindent

\subsection{Ablation studies}

In this section, we perform comprehensive ablation studies to examine the efficacy of each component, supervoxel size as well as the sampling strategy on the final performance. The experiments are conducted in the SemanticKITTI validation set. More ablation studies are put in the supplementary material.

\begin{table}[!t]
\caption{Influence of each component on the final performance.}
\label{each_component_table}
\centering
\vskip -0.3cm
\small{
\begin{tabular}{c|c|c|c|c}
\hline
$\mathcal{L}_{out\_p}$ & $\mathcal{L}_{out\_v}$ & $\mathcal{L}_{aff\_p}$ & $\mathcal{L}_{aff\_v}$ & mIoU \\
\hline
        &         &         &         &  63.1 \\
$\surd$ &         &         &         &  63.4 \\
$\surd$ & $\surd$ &         &         &  64.1 \\
$\surd$ & $\surd$ & $\surd$ &         &  64.7 \\
$\surd$ & $\surd$ & $\surd$ & $\surd$ &  66.4 \\
\hline
\end{tabular}
}
\vspace{-4ex}
\end{table}

\noindent \textbf{Effect of each component.} From Table~\ref{each_component_table}, we have the following observations: 1) Combining both point-to-voxel output distillation and affinity distillation brings the most performance gains. 2) The voxel-level distillation brings more gains than the point-level distillation, suggesting the necessity of introducing the voxel-level mimicking loss. 3) The affinity distillation yields more gains than the output distillation, demonstrating the importance of leveraging the relational knowledge to better capture the structural information. 

\begin{table}[!t]
\caption{Effect of supervoxel size on the performance.}
\label{supervoxel_size_table}
\centering
\vskip -0.3cm
\small{
\begin{tabular}{c|c}
\hline
supervoxel size & mIoU \\
\hline
(60, 30, 4) & 65.3 \\
\hline
(90, 45, 6) & 65.7 \\
\hline
(120, 60, 8) & 66.4 \\
\hline
(180, 90, 12) & 65.6 \\
\hline
(240, 180, 16) & 65.2 \\
\hline
\end{tabular}
}
\vspace{-3ex}
\end{table}

\noindent \textbf{Supervoxel size.} Note that we perform the inter-point and inter-voxel distillation on the sampled supervoxels. Supervoxel size has a non-negligible effect on the efficacy of \algorithmname~since a overly small size will make student learn little from the affinity distillation loss while a large supervoxel size will weaken the learning efficiency of \algorithmname. Here, we keep the number of sampled supervoxels as 4 to remove the effect of this factor. From Table~\ref{supervoxel_size_table}, we can see that setting the supervoxel size to (120, 60, 8) yields the best performance. Remarkably increasing or decreasing the supervoxel size will harm the distillation efficacy.   

%\noindent \textbf{Number of supervoxels.} Remember that we set the number of sampled supervoxels as 4 in our algorithm.

\begin{figure}[t]
 \centering
 \includegraphics[width=0.75\linewidth]{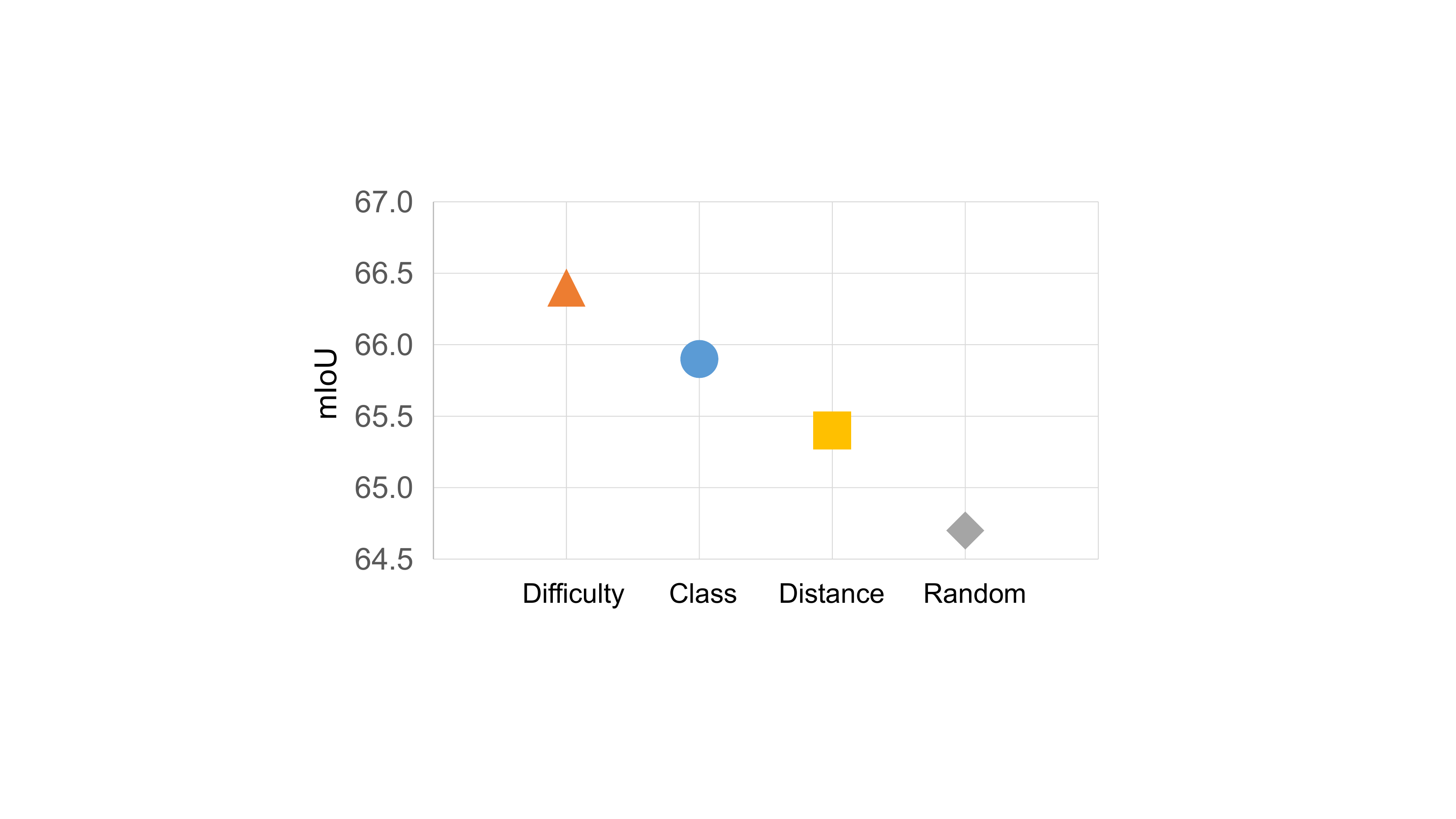}
 \vskip -0.2cm
 \caption{Comparison between different sampling strategies.}
 \centering
 \vskip -0.5cm
 \label{fig:sampling_compare}
\end{figure}

\noindent \textbf{Sampling strategy.} We compare four different sampling strategies, \ie, the original difficulty-aware sampling, distance-aware sampling, category-aware sampling and random sampling. Here, distance-aware sampling is to only more frequently sample distant points while category-aware sampling will more likely sample points belonging to rare classes. From Fig.~\ref{fig:sampling_compare}, it is apparent that difficulty-aware sampling brings more gains than the other three strategies. Specifically, difficulty-aware sampling outperforms both distance-aware and category-aware sampling, suggesting that both distance and categorical awareness are crucial to the distillation effect. The large gap between difficulty-aware sampling and random sampling validates the necessity of difficulty-aware sampling strategy. 

\section{Conclusion}\label{conclusion}

In this paper, we propose a novel point-to-voxel knowledge distillation approach (\algorithmname) tailored for LiDAR semantic segmentation. \algorithmname~is comprised of the point-to-voxel output distillation and affinity distillation. The supervoxel partition and difficulty-aware sampling strategy are further proposed to improve the learning efficiency of affinity distillation. We perform experiments on two LiDAR semantic segmentation benchmarks and show that \algorithmname~significantly outperforms baseline distillation algorithms on distilling Cylinder3D, SPVNAS and MinkowskiNet. The impressive results indicate that there still exists large redundancy in 3D segmentation models and our approach can serve as a strong baseline to compress these cumbersome models.

\noindent \textbf{Acknowledgements.} This work is partially supported under the RIE2020 Industry Alignment Fund - Industry Collaboration Projects (IAF-ICP) Funding Initiative, as well as cash and in-kind contribution from the industry partner(s). 
%Notably, on the nuScenes and SemanticKITTI datasets, our method can achieve roughly \textbf{75\%} MACs reduction and \textbf{2}$\times$ speedup on the competitive Cylinder3D model with negligible performance drop.
%%%%%%%%% REFERENCES
{\small
\bibliographystyle{ieee_fullname}
\bibliography{main}
}

\appendix
\appendixpage
\addappheadtotoc

\section{Quantitative results}

\begin{table}[!t]
\caption{Performance of different algorithms on distilling SPVNAS and MinkowskiNet on SemanticKITTI validation set.}
\vskip -0.3cm
\label{more_spvnas_table}
\centering
\small{
\begin{tabular}{c|c|c}
\hline
Algorithm & mIoU & MACs (G) \\
\hline
\hline
SPVNAS~\cite{tang2020searching} & \bf{63.8} & 118.6 \\
\hline
SPVNAS$\_$0.5$\times$ & 60.4 & \multirow{7}*{29.7}  \\
\cline{1-2}
SPVNAS$\_$0.5$\times$ + KD & 60.6 & ~ \\
\cline{1-2}
SPVNAS$\_$0.5$\times$ + CD & 60.9 & ~ \\
\cline{1-2}
SPVNAS$\_$0.5$\times$ + IFV & 60.8 & ~ \\
\cline{1-2}
SPVNAS$\_$0.5$\times$ + SKD & 61.2 & ~ \\
\cline{1-2}
SPVNAS$\_$0.5$\times$ + KA & 60.7 & ~ \\
\cline{1-2}
\textbf{SPVNAS$\_$0.5$\times$ + \algorithmname~} & \textbf{63.8} & ~ \\
\hline \hline
MinkowskiNet~\cite{choy20194d} & 61.9 & 114.0 \\
\hline
MinkowskiNet$\_$0.5$\times$ & 58.9 & \multirow{7}*{28.5}  \\
\cline{1-2}
MinkowskiNet$\_$0.5$\times$ + KD & 59.2 & ~ \\
\cline{1-2}
MinkowskiNet$\_$0.5$\times$ + CD & 59.6 & ~ \\
\cline{1-2}
MinkowskiNet$\_$0.5$\times$ + IFV & 59.1 & ~ \\
\cline{1-2}
MinkowskiNet$\_$0.5$\times$ + SKD & 59.4 & ~ \\
\cline{1-2}
MinkowskiNet$\_$0.5$\times$ + KA & 59.2 & ~ \\
\cline{1-2}
\textbf{MinkowskiNet$\_$0.5$\times$ + \algorithmname~} & 61.8 & ~ \\
\hline
\end{tabular}
}
\vspace{-3ex}
\end{table}

We provide the complete quantitative results of different algorithms on SPVNAS~\cite{tang2020searching} and MinkowskiNet~\cite{choy20194d} in Table~\ref{more_spvnas_table}. Apparently, \algorithmname~consistently outperforms previous distillation algorithms by a large margin. For instance, on SPVNAS, \algorithmname~can bring 2.6 more points than the SKD method in mIoU. For both models, \algorithmname~can almost mitigate the performance gap between the original network and the pruned model. The encouraging results on SPVNAS and MinkowskiNet convincingly demonstrates the good scalability of \algorithmname.  

\section{Ablation studies}

\noindent \textbf{Loss coefficients.} By comparing each row with the last row in Table~\ref{loss_coefficient_table}, we have the following observations: 1) the loss coefficient of the inter-voxel affinity distillation should be larger than other distillation losses to yield the best distillation effect (row 1 and 2). 2) Exchanging the loss coefficients of the point-based distillation loss and voxel-based distillation loss deteriorates the performance, which means the voxel-based distillation loss guides the point-based loss and is more important (row 5). 3) Slightly increasing the loss coefficients will not significantly affect the overall performance, which demonstrates the robustness of \algorithmname~(row 3 and 4).

\begin{table}[!t]
\caption{Performance of using different loss coefficients for \algorithmname.}
\label{loss_coefficient_table}
\centering
\vskip -0.3cm
\small{
\begin{tabular}{c|c|c|c|c}
\hline
$\alpha_{1}$ & $\alpha_{2}$ & $\beta_{1}$ & $\beta_{2}$ & mIoU \\
\hline
0.1 & 0.1 & 0.1 & 0.1 &  65.2 \\
0.25 & 0.25 & 0.25 & 0.25 &  65.5 \\
0.15 & 0.15 & 0.15 & 0.25 &  66.2 \\
0.1 & 0.2 & 0.2 & 0.25 & 66.3 \\
0.15 & 0.1 & 0.25 & 0.15 &  65.4 \\
0.1 & 0.15 & 0.15 & 0.25 &  66.4 \\
\hline
\end{tabular}
}
\vspace{-3ex}
\end{table}

\noindent \textbf{Performance sensitiveness to $f_{class}$.} We conduct experiments on examining the effect of $f_{class}$. We rewrite the $f_{class}$ to be $f_{class} = \alpha \exp{(\beta N_{minor})}+1$. Then, we randomly choose $\alpha$ from \{3, 4, 5, 6\} and $\beta$ from \{ -1, -2, -3\}, and compare the performance of different combinations. Experimental results reveal that the final performance of \algorithmname~on Cylinder3D\_0.5$\times$ ranges from 66.2 to 66.4. The small fluctuations in distillation performance indicates that \algorithmname~is not very sensitive to $f_{class}$.

\begin{table}[!t]
\caption{Influence of each component on the final performance.}
\label{each_component_table_more}
\centering
\vskip -0.3cm
\small{
\begin{tabular}{c|c|c|c|c}
\hline
$\mathcal{L}_{out\_p}$ & $\mathcal{L}_{out\_v}$ & $\mathcal{L}_{aff\_p}$ & $\mathcal{L}_{aff\_v}$ & mIoU \\
\hline
        &         &         &         &  63.1 \\
$\surd$ &         &         &         &  63.4 \\
        & $\surd$ &         &         &  63.7 \\
        &         & $\surd$ &         &  63.6 \\
        &         &         & $\surd$ &  64.5 \\
\hline
\end{tabular}
}
\vspace{-5ex}
\end{table}

\noindent \textbf{The influence of each component of \algorithmname.} Detailed performance of each component is summarized in Table~\ref{each_component_table_more}. The voxel-based loss term indeed has larger impacts on the final performance. One potential reason is that the voxel representation provides richer structural information of the environment as it aggregates the information of all points within a voxel.

\noindent \textbf{Broader impact of \algorithmname.} We apply \algorithmname~to SemanticKITTI multi-scan segmentation tasks and observe \textbf{4.2$\%$} performance improvement on the Cylinder3D\_0.5$\times$ backbone. The resulting model ranks 3rd on the SemanticKITTI multi-scan competition~\footnote{https://competitions.codalab.org/competitions/20331\#results (multi-scan competition) till 2021-12-1 00:00 Pacific Time, and our method is termed PV-KD}.

\noindent \textbf{Performance w.r.t the distances of objects.} We repartition SemanticKITTI according to the position of the cars. Cars in the training set are relatively close to the origin ($\leq$ 20 m) while cars in the validation set are relatively far away from the origin ($>$ 20 m). We apply \algorithmname~to distill the Cylinder3D model on the newly divided dataset. Experimental results show that \algorithmname~can still bring \textbf{4.7}\% to the Cylinder3D\_0.5$\times$ model on cars (91.3\% v.s. 95.6\%).

\begin{table*}[!t]
\caption{Loss coefficients of different distillation methods.}
\label{baseline_loss_coefficient_table}
\centering
\vskip -0.3cm
\small{
\begin{tabular}{c|c|c|c|c|c|c|c|c|c|c|c|c}
\hline
\multirow{2}*{Model} & \multirow{2}*{KD} & \multirow{2}*{IFV} & \multicolumn{2}{c|}{SKD} & \multicolumn{2}{c|}{CD} & \multicolumn{2}{c|}{KA} & \multicolumn{4}{c}{\algorithmname} \\
\cline{4-13}
~ &  & & $\lambda_{pi}$ & $\lambda_{pa}$ & $\lambda_{fea}$ & $\lambda_{score}$ & $\lambda_{ada}$ & $\lambda_{aff}$ & $\alpha_{1}$ & $\alpha_{2}$ & $\beta_{1}$ & $\beta_{2}$ \\
\hline
Cylinder3D & 0.2 & 0.2 & 0.15 & 0.2 & 0.1 & 0.3 & 0.2 & 0.15 & 0.1 & 0.15 & 0.15 & 0.25\\
\hline
SPVNAS & 0.1 & 0.3 & 0.15 & 0.15 & 0.05 & 0.1 & 0.2 & 0.1 & 0.1 & 0.1 & 0.1 & 0.15 \\
\hline
MinkowskiNet & 0.2 & 0.2 & 0.1 & 0.2 & 0.05 & 0.1 & 0.15 & 0.2 & 0.1 & 0.1 & 0.15 & 0.2 \\
\hline
\end{tabular}
}
\vspace{-3ex}
\end{table*}

\section{Elaborated implementation details}

For the baseline knowledge distillation approaches, the value of each loss coefficient is provided in Table~\ref{baseline_loss_coefficient_table}. Since training a single model from scratch may take one more week, we resort to loading the pre-trained weights to accelerate the training process. In this condition, the overall training duration will be shortened to three days. Note that all methods adopt this strategy to ensure fair comparison. The latency is recorded using a single GPU (NVIDIA Tesla PG503-216 GV100) and the final value of latency is obtained after averaging the latency of 100 samples. The training protocol for SPVNAS and MinkowskiNet is exactly the same as their open-sourced codes\footnote{https://github.com/mit-han-lab/spvnas}. Finetuning denotes retraining the trained model for 10 more epochs with the learning rate being 2e-4.

\noindent \textbf{MACs calculation:} Since sparse convolution merely operates on the non-zero positions and different input point cloud sequence has different non-zero patterns, we first estimate the average kernel map size following~\cite{tang2020searching} for each layer and then use the following equation to compute the FLOPs of each layer: $FLOPs = N \times K_{s} \times C_{in} \times C_{out}$, where $K_{s}$ is the size of kernel map, $N$ is the number of points, $C_{in}$ is the number of input channels, $C_{out}$ is the number of output channels.

%\section{More experiments}

\section{Qualitative results}

\begin{figure*}[t]
 \centering
 \includegraphics[width=1.0\linewidth]{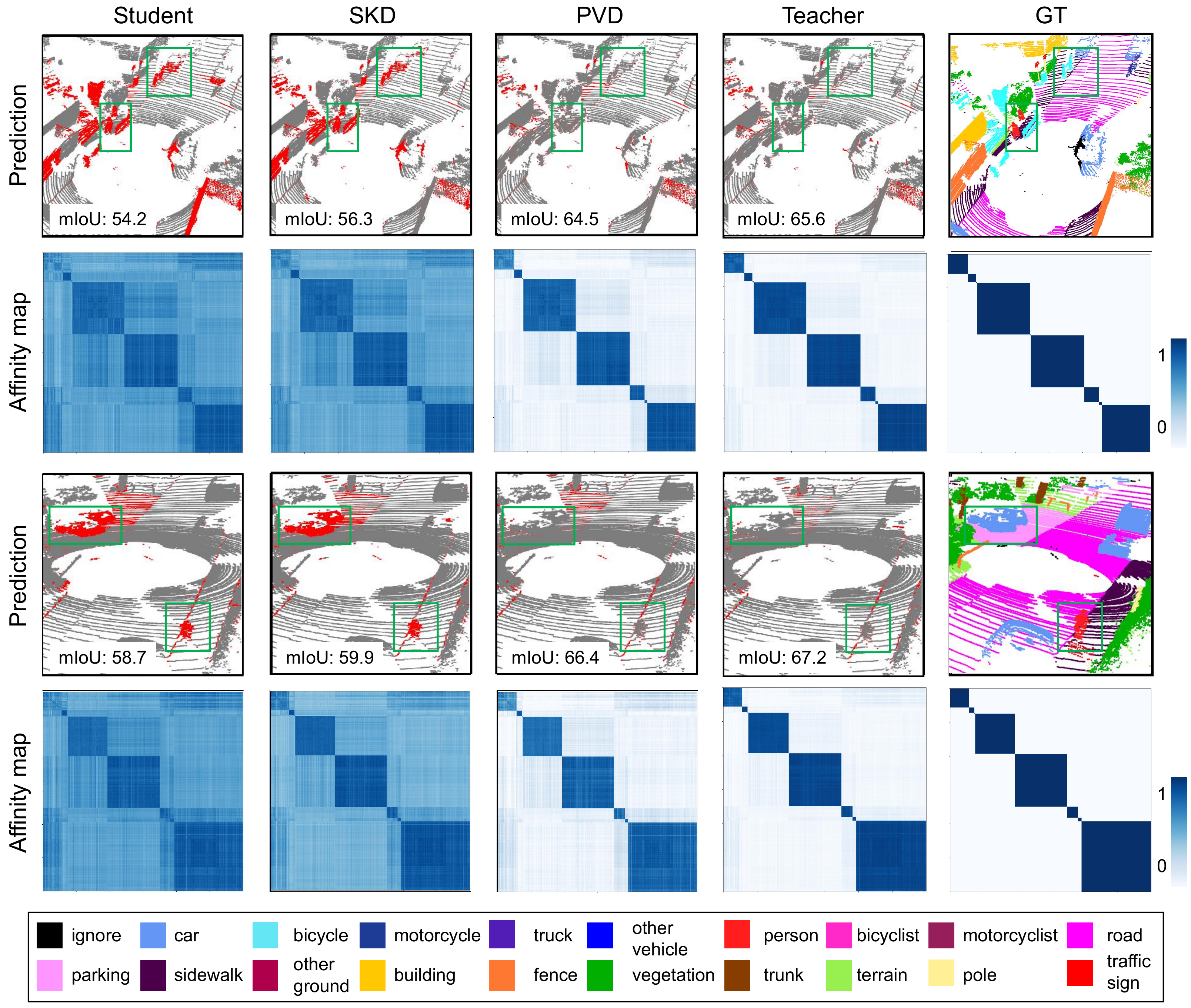}
 \vskip -0.3cm
 \caption{Visual comparison of different methods on the SemanticKITTI validation set. Here, the ground-truth for the inter-voxel affinity map is the ideal map where the intra-class similarity score is 1 and inter-class similarity score is 0.}
 \centering
 \vskip -0.6cm
 \label{fig:more_visual_compare}
\end{figure*}

We provide more visual comparisons of \algorithmname~with previous distillation algorithms in Fig.~\ref{fig:more_visual_compare}. Compared with the rival SKD approach, our \algorithmname~can significantly improve the prediction of the student model. The prediction errors of \algorithmname~on those minority classes, \eg, person and bicycle, are much smaller than those of SKD. Besides, on objects that are faraway from the origin, \eg, the car highlighted by the green rectangle, \algorithmname~also yields more accurate predictions than SKD. And \algorithmname~has lower inter-class similarity and higher intra-class similarity than SKD. The aforementioned results explicitly demonstrate the efficacy of \algorithmname~in distilling structural knowledge from the teacher model to the student.

\end{document}